\documentclass{article}

\usepackage[preprint]{corl_2026}

\usepackage{amsmath,amssymb}
\usepackage{booktabs}
\usepackage{multirow}
\usepackage{graphicx}
\usepackage{xcolor}
\usepackage{url}

\usepackage{subcaption}
\newcommand{\statfigwidth}{0.65\textwidth}
\usepackage{wrapfig} 

\usepackage{tabularx}
\usepackage{array}
\newcolumntype{Y}{>{\centering\arraybackslash}X}

\newcommand{\heldset}{\mathcal{F}_h}
\newcommand{\fixedset}{\mathcal{F}_f}
\newcommand{\heldobj}{\mathcal{O}_h}
\newcommand{\fixedobj}{\mathcal{O}_f}

\title{Assembling Two Parts in One Hand}

\author{
  Liuao Pei\textsuperscript{1,2,*}\quad
  Tianyue Wu\textsuperscript{1,2,*}\quad
  Hui Zhang\textsuperscript{4}\quad
  Ping Luo\textsuperscript{1,\textdagger}\quad
  Jie Song\textsuperscript{2,3,\textdagger}\\[4pt]
  \normalfont\small\textsuperscript{1}The University of Hong Kong\\
  \normalfont\small\textsuperscript{2}The Hong Kong University of Science and Technology (Guangzhou)\\
  \normalfont\small\textsuperscript{3}The Hong Kong University of Science and Technology\\
  \normalfont\small\textsuperscript{4}ETH Zurich\\[4pt]
  \normalfont\footnotesize\textsuperscript{*}Equal contribution. The first two authors are listed in alphabetical order.\\
  \normalfont\footnotesize\textsuperscript{\textdagger}Corresponding authors.
}

\begin{document}
\nolinenumbers
\maketitle

\begin{figure}[h]
    \centering
    \includegraphics[width=\linewidth]{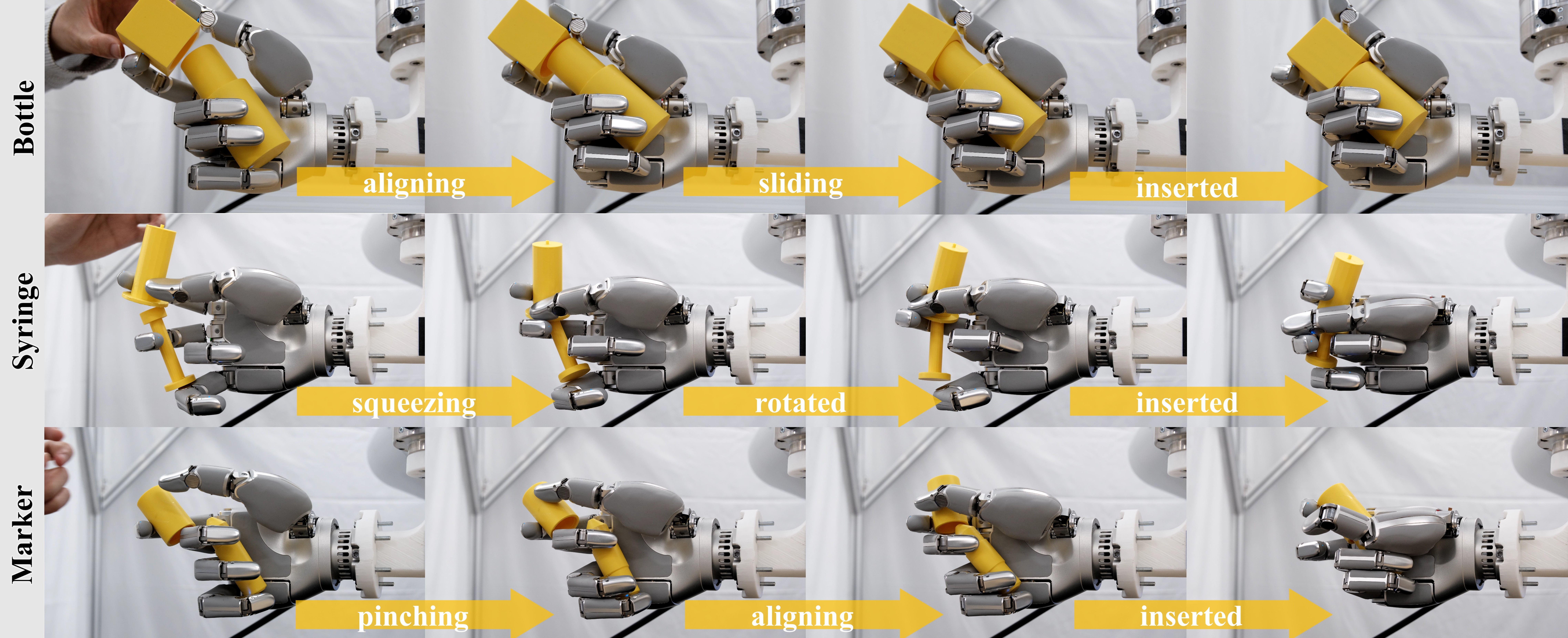}
    \caption{\textbf{Real-world in-hand assembly.} A single dexterous hand mates two
    objects using fine finger coordination without a second arm or external fixture, shown here on the Bottle, Syringe, and Marker tasks.}
    \label{fig:teaser}
    \vspace{-0.2cm}
\end{figure}

\begin{abstract}
A hallmark of human dexterity is the cooperative use of fingers, where different fingers take on distinct yet coordinated roles to accomplish fine manipulation, such as capping a pen with the hand that holds it. We study this finger-level coordination through \emph{in-hand assembly}: mating two rigid objects within a single dexterous hand, with no second arm and no fixture. We present a reinforcement learning formulation to solve this problem in a unified framework, which is driven by a goal relative pose between the two parts. Finger coordination is shaped by a function-based auxiliary reward and regularized toward a single human reference pose, while domain randomization and a fusion of historical proprioception and object observation confer robustness to occlusion-induced estimation noise. The same recipe solves three different assembly tasks (Bottle, Syringe, and Marker). Trained purely in simulation, the policies transfer zero-shot to hardware with a single camera, demonstrating robustness to state-estimation errors caused by occlusion. Our experiments also reveal that in-hand assembly places demands on hand morphology and can serve as a benchmark for modern robotic hand systems.
Videos and code are available at \href{https://ltbgbird.github.io/in-hand-assembly-page/}{this URL}.
\end{abstract}


\vspace{-0.2cm}
\section{Introduction}
\label{sec:intro}
\vspace{-0.2cm}

While animals can also perform basic manipulation behaviors with their limbs or digits, such as pushing or pulling objects, the ability to execute complex manipulation through coordinated finger interactions is most refined in human hands~\cite{marzke1997precision,feix2015estimating}. For instance, humans squeeze a syringe with one hand, cap a pen with the hand that holds it, or press a lid back onto a bottle held in the same hand.
This cooperative use of individual fingers is common in everyday human manipulation, whereas current work on dexterous manipulation remains limited to relatively simple or repetitive finger motions and contact modes, as exemplified by pick-and-place tasks \cite{wang2022dexgraspnet,qin2023dexpoint,zhang2025robustdexgrasp,lin2025sim} and in-hand reorientation \cite{chen2022system,qi2023hand,liu2025dexndm}. These capabilities have yet to fully exploit the dexterity of modern robotic hands.

To move beyond these limited settings, we study tasks that more directly reflect and exploit the coordinative dexterity of human hands. Specifically, we focus on \emph{in-hand assembly}: tasks in which a single multi-fingered hand simultaneously manipulates two objects to mate them and operates the assembly under articulation constraints (Fig. \ref{fig:teaser}). 
In-hand assembly tasks, however, are inherently challenging in several aspects.
First, the tasks are contact-rich and control-intensive, requiring forceful yet precise coordination across four or five fingers. Such intricate multi-finger motions are difficult to obtain through teleoperation \cite{qin2023anyteleop} or finger-by-finger kinesthetic demonstration \cite{chen2025dexforce}.
Second, assembly tasks exhibit diversity in object geometry, mating mechanisms, and target motions. Together with the large number of degrees of freedom in the control space, this often leads to task-specific formulations and optimization objectives, posing a major scalability challenge.
Third, assembly requires precise alignment, while occlusion-induced errors in object state estimation pose significant challenges to control robustness.

To address these challenges, we develop a policy learning approach based on reinforcement learning (RL) in simulation to automatically explore solutions.
From the perspective of relative object state goal reaching, we abstract a shared task formulation that captures reusable structure across assembly tasks, with different roles assigned to different fingers, such as object translocation and compliant support, triggered by a function-relevant reward.
We further propose reference snapshots from human manipulation to constrain exploration. 
To enable robust closed-loop execution under occlusion, we apply observation domain randomization during training, encouraging the policy to implicitly calibrate noisy estimates through proprioceptive feedback.

Our contributions are threefold:
(1) \textbf{In-hand assembly as a dexterity benchmark}. We introduce in-hand assembly as a challenging dexterous manipulation setting. By requiring one hand to coordinate multiple fingers across two interacting objects, it provides a task family for evaluating coordinative dexterity in robotic manipulation systems.
(2) \textbf{Systematic solution}. We present a systematic solution for in-hand assembly that integrates simulation-based RL, finger function rewards, and human kinematic priors, which is unified across tasks. To our knowledge, these components enable the first fixture-free assembly demonstration through a general-purpose anthropomorphic hand.
(3) \textbf{Real-world effectiveness}. The system performs real-world in-hand assembly using only stereo vision from a single camera and historical proprioception, demonstrating effectiveness under state-estimation noise and unexpected disturbance.

\vspace{-0.4cm}
\section{Related Work}
\label{sec:related}

\vspace{-0.3cm}
\subsection{Dexterous Manipulation with Multi-Fingered Hands}
\vspace{-0.2cm}
Most work on multi-fingered hands centers on grasping, either synthesizing large-scale datasets~\cite{wang2022dexgraspnet, zhang2024graspxl, ye2025dex1b} or training policies to grasp novel objects~\cite{lum2024dextrahg, huang2025fungrasp, chen2025clutterdexgrasp, zhong2026dexgraspvla}. Such grasps often reduce the hand to enveloping a single object with all fingers, which a low-DoF gripper can also achieve~\cite{fang2023anygrasp}. A smaller body of work grasps several objects with one hand~\cite{seqmultigrasp, seqgrasp, lu2025grasping}, but partitions the fingers into subsets that each enclose a separate object, acting as independent grippers that co-hold the objects rather than coordinating them, leaving the coordinative dexterity limited. Another line uses sim-to-real RL~\cite{openai2019rubik,andrychowicz2020learning} for in-hand manipulation of a single object, typically on easy-to-formulate skills such as in-hand rotation~\cite{chen2022system, qi2023hand, liu2025dexndm, yang2024anyrotate,yin2023rotating, yin2025dexteritygen} or translation~\cite{yin2025learning}. Others learn from teleoperated demonstrations~\cite{chen2025dexforce, cheng2024open, yang2024ace, lin2025learning} or human motion retargeting~\cite{li2025maniptrans,liu2025dextrack}, but the embodiment gap between human and robot hands hinders agile, highly dexterous skills with high precision. Overall, existing systems still do not capture the human-like dexterity of precisely coordinating multiple objects within a single hand. In contrast, we use multi-object in-hand assembly to expose this gap and probe the boundary of machine dexterity with modern multi-fingered hands.
\vspace{-0.2cm}
\subsection{Robotic Assembly}
\vspace{-0.2cm}
Assembly is a long-standing problem in robotic manipulation, with insertion (peg-in-hole) as its canonical instance. Current learning-based methods show that sim-to-real RL can acquire robust contact-rich insertion policies~\cite{narang2022factory, tang2023industreal, tang2024automate,tian2025fabrica}. Almost all of this work, however, uses a parallel-jaw gripper on a robot arm: the gripper rigidly clamps one part while the other is held by a fixture or a second arm. Lee et al.~\cite{lee2022peg} perform keyhole peg-in-hole with a dual-arm system equipped with robotic hands in which the fingers only grasp and pre-rotate each part. More broadly, two-object contact-rich skills with multi-fingered hands are performed bimanually, e.g., twisting lids off bottles with one hand stabilizing and the other turning~\cite{lin2024twisting}, and target disassembly rather than the harder mating of parts. Triyonoputro et al.~\cite{triyonoputro2018double} demonstrate in-hand assembly with a specialized double-jaw mechanism. We instead study fixture-free assembly through coordinated fingers of a general-purpose anthropomorphic hand, providing a benchmark for its intrinsic dexterity.
\begin{figure}[t]
    \centering
\includegraphics[width=1\linewidth]{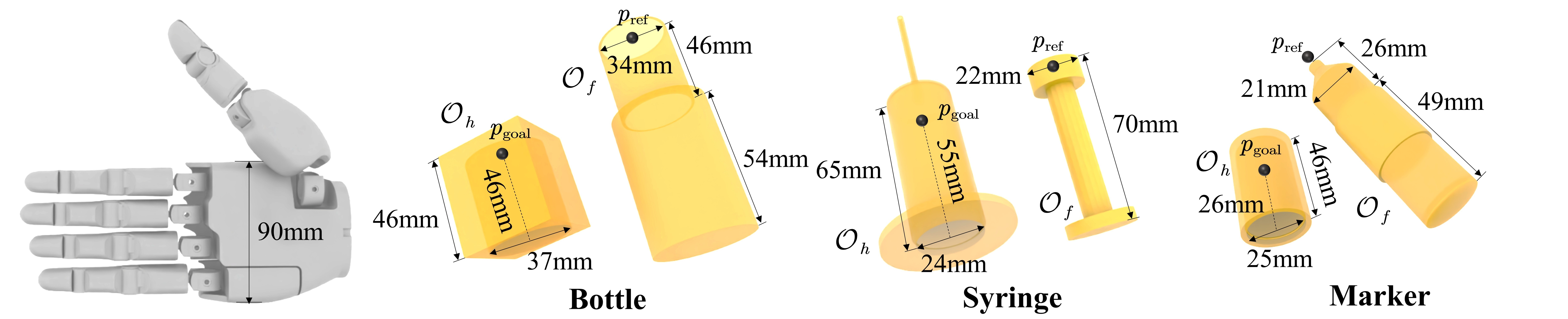}
    \caption{\textbf{Task illustration with key geometric parameters.} This figure includes the semi-transparent parts of three task instances, key geometric parameters, and the positions of the reference point $p_{\text{ref}}$ and goal point $p_{\text{goal}}$ to define a goal-reaching reward (see \S\ref{sec:method:reward}).}
    \label{fig:task_instances}
    \vspace{-0.5cm}
\end{figure}

\vspace{-0.2cm}
\section{Method}
\label{sec:method}
\vspace{-0.2cm}
We formulate \emph{in-hand assembly} as a 
reinforcement learning problem (\S\ref{sec:method:problem}), and introduce three tasks for this setting (\S\ref{sec:method:instances}). Based on this, we design our unified observation \& action spaces (\S\ref{sec:method:obs}), and reward structure (\S\ref{sec:method:reward}). We also involve human manipulation snapshots for initialization (\S\ref{sec:method:init}), and apply domain randomization during training to improve robustness (Appx. \S\ref{sec:appendxA}). The  simulation setup especially for contact-rich assembly \cite{narang2022factory}, RL implementation details, and policy architectures are postponed to Appx. \ref{sec:appendxA}.

\vspace{-0.2cm}
\subsection{Problem Formulation}
\label{sec:method:problem}
\vspace{-0.2cm}
Let $\heldobj$ and $\fixedobj$ be the two rigid bodies to be mated, and let
$\mathcal{H}$ be a \emph{single} dexterous hand with $n_q$ revolute joints. At every control step, the hand must drive $\heldobj$ into a \emph{goal relative pose} with respect to $\fixedobj$ that satisfies a task-specific geometric mating constraint.  The names (held versus fixed) follow
the assembly literature \cite{tang2023industreal}, whereas no second arm, table fixture, or vise is available; both objects are supported by the same hand. 

We partition $\mathcal{H}$'s fingers into two role sets and regularize their motion using separate reward components (see \S\ref{sec:method:reward}) that softly constrain which fingers contact which object. The resulting behavioral specialization (reported in \S\ref{sec:experiments}) is: the held set $\heldset = \{\text{thumb}, \text{index}\}$ manipulates $\heldobj$ at its top and provides the fine alignment and goal-pose-reaching motion. The fixed set $\fixedset = \{\text{middle}, \text{ring}, \text{little}\}$ cages $\fixedobj$ and resists the reaction wrenches generated by $\heldset$, while also assisting with the alignment and assembly, e.g., the middle and ring fingers might fine-tune the pose of $\fixedobj$, and the pinky finger can assist with insertion.

Casting this as a partially observable Markov decision process (POMDP)
$(\mathcal{S}, \mathcal{A}, \mathcal{O}, T, R, \gamma)$, we seek a recurrent policy
$\pi_\theta(a_t \mid o_{\le t})$ that maximizes
$\mathbb{E}_\pi\!\left[\sum_{t=0}^{T_\text{max}} \gamma^t R_t\right]$ subject
to joint-limit and contact-feasibility constraints.

\vspace{-0.2cm}
\subsection{Task Instances}
\label{sec:method:instances}
\vspace{-0.2cm}

We have instantiated in-hand assembly into three tasks: (1) Bottle cap and bottle assembly (Bottle); (2) Syringe insertion (Syringe); (3) Marker cap fitting (Marker), as shown in Fig. \ref{fig:task_instances}.  The primary differences among these three assembly tasks lie in the parts’ geometries, assembly hand pose, and tolerances. For instance, the parts in the Bottle task are relatively large, which can lead to unstable grasping, and adjusting the bottle’s orientation (often necessary) requires the involvement of each finger's motion. In the Syringe task, the plunger requires contact with the back of the ring finger and the involvement of the pinky finger for insertion, while, in the Marker task, the hand causes extensive occlusion, placing demands on the robustness of the state estimator and the policy against estimation errors. The assembly types for all three tasks are similar to those described in previous assembly literature \cite{tang2023industreal,lee2022peg}, specifically plug-in assembly. In the future, we plan to expand to other types of assembly, such as threading and friction-fit assemblies; however, these will require more accurate simulation \cite{hsieh2025learning}.

\vspace{-0.2cm}
\subsection{Observation and Action Spaces}
\label{sec:method:obs}
\vspace{-0.2cm}
\paragraph{Observation.} The policy reads a 34-dimensional vector
\begin{equation}
    o_t \;=\; \bigl[\,
        p^{(h)}_t \,\big|\, \hat{z}^{(h)}_t \,\big|\,
        p^{(f)}_t \,\big|\, \hat{z}^{(f)}_t \,\big|\,
        q_t 
    \,\bigr]
    \;\in\; \mathbb{R}^{3+3+3+3+22},
    \label{eq:obs}
\end{equation}
where $p^{(\cdot)} \in \mathbb{R}^3$ is the object centroid in the hand-base
frame, $\hat{z}^{(\cdot)} \in S^2$ is the unit vector representing
the object's body-frame $z$-axis, which is the object's axis of rotational symmetry, in the world coordinates,
and $q_t \in \mathbb{R}^{22}$ is the vector of measured joint angles. If we randomize hand tilt during training, the observation space is extended to include $\phi$ and $\theta$, the roll and pitch angles of the hand in the world frame.

We note that the observation space consists solely of the incomplete state of the object derived from visual state estimation and the robot’s proprioception; it does not include any higher-order quantities such as joint velocities or contact force signals. Since most of the components being assembled are rotationally symmetric about their body z-axis, we omit the estimation of this degree of freedom. The policy is represented as a recurrent neural network (RNN) to tackle the partial observability.

\paragraph{Action.} The action $a_t \in [-1, 1]^{22}$ is a normalized
joint-position-target delta applied to a stiff implicit Proportional-Derivative (PD) controller:
\begin{equation}
    q^\text{target}_{t+1} \;=\; \text{clip}\!\left( q_t + \alpha\, a_t,\; q_\text{min},\; q_\text{max} \right),
    \quad \alpha = 0.1\ \text{rad},
    \label{eq:action}
\end{equation}
where $q_t$ is the current PD joint-position target, and $q_\text{min}$ and $q_\text{max}$ represent the lower and upper joint limits, respectively. We control at 15 Hz, and this action structure results in naturally smooth absolute joint angles.

\paragraph{Real-World Object Pose Tracking.} 
At deployment, we recover the 6-DoF poses of $\heldobj$ and $\fixedobj$ from an RGB-D stream with FoundationPose~\cite{wen2024foundationpose}, which registers each object mesh against the first frame and refines the pose per frame thereafter. Each pose is then reduced to the $(p,\hat{z})$ form of Eq.~(\ref{eq:obs}); a lightweight depth-consistency gate rejects per-frame estimates whose predicted depth disagrees with the measured depth at the projected center, which prevents the tracker from locking onto the manipulating hand when it occludes the object. More details can be found in Appx. \ref{sec:method:deploy}.

\vspace{-0.2cm}
\subsection{Goal-Reaching with Auxiliary Rewards}
\label{sec:method:reward}
\vspace{-0.2cm}

\paragraph{Goal-reaching reward.}  The main reward signals come from:
\vspace{-0.1cm}
\begin{equation}
    R^\text{goal}_t \;=\; 5 \cdot (\underbrace{\exp(-60e_{xy}-60e_z-5e_{\theta})}_{\in [0,1]} + \underbrace{\exp(-200e_{xy}-120e_z-30e_{\theta})}_{\in [0,1]}), 
    \label{eq:reward}
\end{equation}
where $e_{xy}$ and $e_{z}$ are the xy-plane and z-axis distances between the reference point and the goal point illustrated by Fig. \ref{fig:task_instances} in $\fixedobj$'s local frame (in meters), and $e_{\theta} = 1 - z_{\text{held}}\cdot z_{\text{fixed}}$, with  $z_{\text{held}}$ and   $z_{\text{fixed}}$ being the normalized direction vectors of the body $z$-axes of $\heldobj$ and $\fixedobj$, respectively. This reward describes the achievement of relative spatial relationship between the two parts, which is the objective of the assembly task. We set up two additive terms to provide goal-reaching motivations at different scales.

\paragraph{Auxiliary reward.} Active throughout the episode, it is a
multiplicative combination of three normalized quality signals:
\vspace{-0.1cm}
\begin{equation}
    R^\text{aux}_t \;=\; 
    \underbrace{\rho^\text{pinch}_t}_{\in [0,1]} \;\cdot\;
    \underbrace{\rho^\text{grasp}_t}_{\in [0,1]} \;\cdot\;
    \underbrace{\rho^\text{pose}_t}_{\in [0,1]}.
    \label{eq:track}
\end{equation}
$\rho^\text{pinch}_t$ counts how many of \{thumb, index\} are in contact
with $\heldobj$; $\rho^\text{grasp}_t = \exp(-20 \cdot
\max_{i \in \fixedset} d_i)$ rewards $\fixedset$ caging $\fixedobj$;
$\rho^\text{pose}_t = \exp(-\|q_t - q_0\|_M^2 / 12)$ constrains drift from a
nominal grasp pose $q_0$ under a per-joint mask $M$. These rewards reflect an empirically motivated organization of the fingers: the thumb and index finger work together to manipulate the held object (a “pinch”), while the middle, ring, and pinky fingers manipulate the fixed object (a “grasp”).  The pose constraint is inspired by prior work that learns from human video, where extracted references guide wrist motion and grasping pose \cite{chen2025vividex,lum2025crossing}. For in-hand assembly, however, we apply such references only as single-frame snapshots: the fine, intricate finger motion of in-hand manipulation is susceptible to the embodiment gap and extraction noise, so multi-frame references offer little usable guidance for finger motion. The reward scheme leaves the fine-grained finger motion to  the goal-reaching reward.

\begin{figure}[h]
    \centering
    \vspace{-0.2cm}
\includegraphics[width=0.95\linewidth]{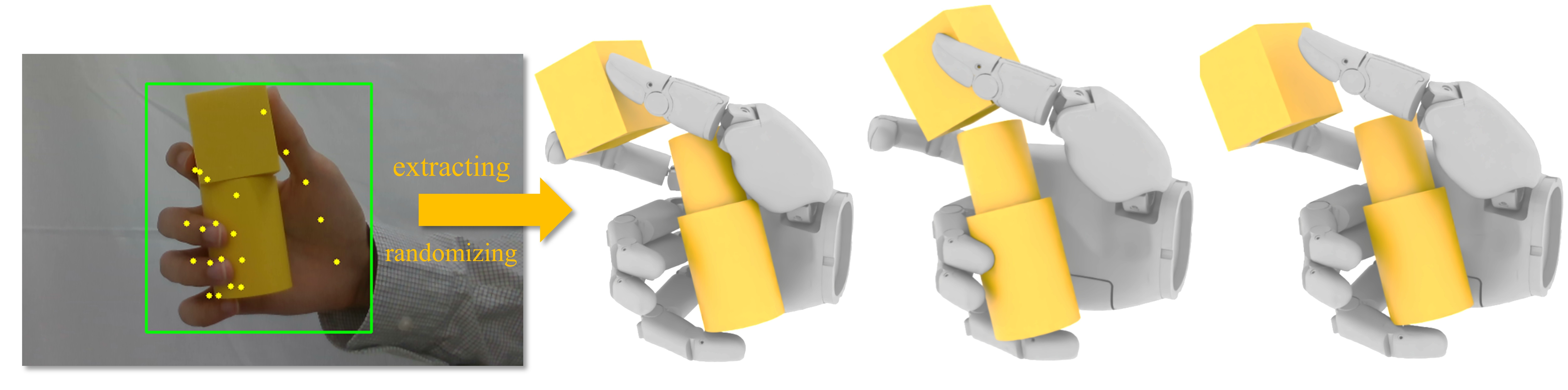}
    \caption{\textbf{State initialization from a human reference snapshot.} On the left is the human reference snapshot for the Bottle task; the three simulated states on the right are initial state examples generated based on the reference hand pose. Initialization for other tasks is postponed to Fig. \ref{fig:appx_init}}
    \label{fig:init}
    \vspace{-0.5cm}
\end{figure}

\subsection{Snapshot as Initialization}
\vspace{-0.2cm}
\label{sec:method:init}
Since the hand is anchored and cannot grasp from a tabletop, every episode must start with the objects already in hand. From a single human-demonstration snapshot of the target task, we geometrically retarget~\cite{qin2023anyteleop} to generate a robotic hand state based on the human hand pose with a 30\% reduction of joint angles of the index finger to widen the opening formed by the index finger and thumb. We then apply the state estimator to extract the objects' poses. The two sources are independent, so directly composing them usually produces hand–object penetration. We resolve this and inject randomization in initial states with a settling phase: holding the hand as a fixed collider, we apply a small random perturbation force and torque to each object and step the simulator for a few steps; the perturbation is then removed, and the episode begins from the resulting state. This procedure settles the objects into randomized configurations, crucially broadening the state coverage seen during training. A more detailed randomization procedure is postponed to Appx. \ref{sec:appendxA}. The objects' gravity is removed for the first few steps in an episode to prevent them from quickly falling. In this way, the policies learn to establish contacts with the objects in these warm-up steps.

\vspace{-0.3cm}
\section{Results}
\vspace{-0.3cm}
\label{sec:experiments}
\subsection{Experimental Setup}
\vspace{-0.2cm}
We use a 22-degree-of-freedom (DoF) Sharpa Wave Hand \cite{sharpawave} in most of the simulation and real-world experiments, which is fixed on a Franka Research 3 arm \cite{fr3}.
The simulation experiments are conducted in IsaacSim \cite{isaacsim}. 
We rely on a single RealSense D435 camera \cite{realsensed435} to perform state estimation. We send control commands to the robot via a Linux workstation with an RTX-3090 Graphics Processing Unit (GPU). See Appx. \ref{sec:method:deploy} for more details of real-world deployment.
 
\vspace{-0.2cm}
\subsection{In-Hand Assembly as a Hand Morphology Benchmark}
\begin{figure}[t]
    \centering
\includegraphics[width=1.03\linewidth]{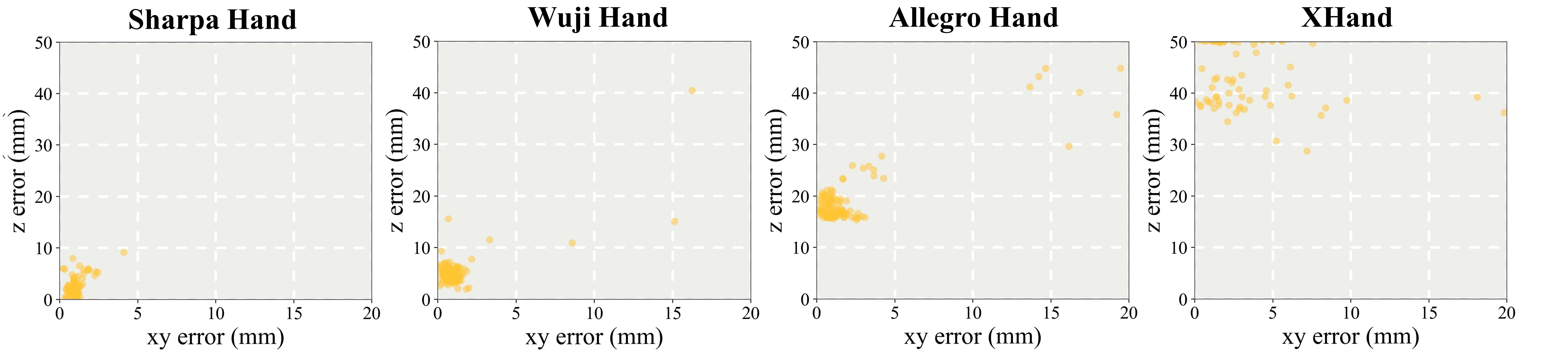}
    \caption{\textbf{Goal-reaching errors for different hand morphologies.} The yellow points in the figures are error samples with a total number of 128.}
    \label{fig:different_hands}
    \vspace{-0.3cm}
\end{figure}

\vspace{-0.2cm}
Intuitively, in-hand assembly places stronger demands on the morphology of the robotic hand than prior in-hand manipulation tasks, which can be accomplished by a wider variety of hands~\cite{liu2025dexndm}. We test this hypothesis by transferring the same pipeline to different hands in simulation. Beyond the Sharpa Hand used in our main experiments, we evaluate the Wuji Hand \cite{wujihand}, the Allegro Hand \cite{allegrohand}, and the XHand \cite{xhand}. Fig.~\ref{fig:different_hands} reports the distribution of reference-point errors on the Syringe task. These errors are obtained from trials that do not terminate early.

From the results, we can tell that the Sharpa Hand and Wuji Hand perform best, owing to their human-like dimensions, high DoF, and wide joint ranges. The Allegro Hand has only four fingers and is larger than a typical human hand, but each finger has many degrees of freedom and a wide joint range, enabling it to align objects in the xy-plane; the lack of a fifth finger, however, prevents the deep insertion required along the z-axis. The XHand is close to human-hand size but has fewer degrees of freedom—most of its fingers lack abduction/adduction joints—and it fails to insert the objects, and the objects cannot be aligned in the xy-plane in many cases.

\vspace{-0.2cm}
\subsection{Simulation Ablation}
\begin{wrapfigure}[19]{r}{0.65\linewidth}
        \centering
        \includegraphics[width=\linewidth]{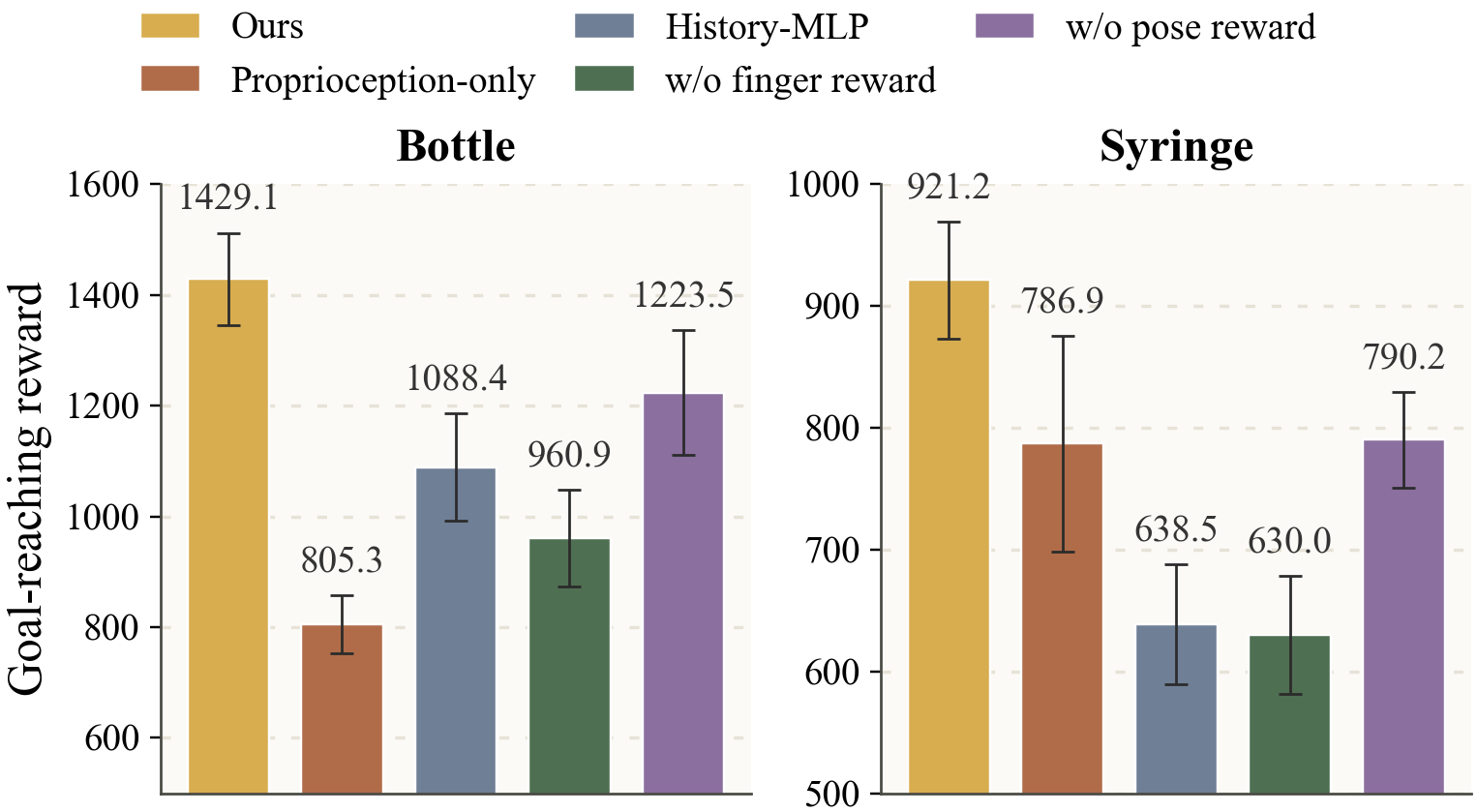}
        \caption{\textbf{Ablation study.} We compare our full policy against:
        (1) \emph{proprioception-only}, which omits object state estimation;
        (2) \emph{history MLP}, which replaces the LSTM with an MLP over 8 frames
        of input--output history;
        (3) \emph{w/o finger reward}, trained without the finger-function contact
        reward $\rho^\text{pinch}\!\cdot\!\rho^\text{grasp}$; and
        (4) \emph{w/o pose reward}, trained without the reference-pose reward
        $\rho^\text{pose}$.}
        \label{fig:ablation}
        \vspace{-0.2cm}
\end{wrapfigure}

\vspace{-0.1cm}
\label{sec:ablation}
We compare our policy against the following variants of our method: 1) A proprioception-only policy. This is a recurrent policy that does not incorporate object state estimation. 2) A historical input-output MLP policy (history MLP) \cite{li2025reinforcement}: instead of using an RNN, this policy uses 8 frames of historical policy input-output  data as policy observations, implemented as an MLP. 3) A policy that does not use the finger-function-based contact reward $\rho^\text{pinch} \cdot \rho^\text{grasp}$ during training (w/o finger reward). 4) A policy that does not use the reference-pose-based reward $\rho^\text{pose}$ during training (w/o pose reward). The results in Fig. \ref{fig:ablation} show the performance metric calculated using the cumulative goal-reaching reward across episodes of the models obtained at epoch 600 across 3 independent training runs. Episodes that end within 0.5s of the warmup are not included in the calculation, as they are considered to start in a poor or infeasible state. The corresponding Marker results are reported in Appendix~\ref{sec:marker_analysis}.

The results show that different tasks demonstrate different requirements for vision, and removing vision generally reduces task performance. For instance, removing vision significantly impacts the performance of the Bottle task. Prior work indicates that certain in-hand manipulation skills can be performed well with proprioception alone~\cite{qi2023hand,liu2025dexndm}. Assembly, however, sometimes requires the policy to have ready access to the spatial relationship between the two objects, so having this information makes learning easier and makes it feasible to manipulate the two objects in precise relative alignment based on explicit object states. We also find that LSTM achieves better performance than stacking historical information within a fixed sample budget, implying that recurrent policies yield a better solution to the POMDP. Meanwhile, we observe that removing some components of the reward affects learning efficiency and the policies' rollout behavior at convergence. This is because the system has a high degree of operational freedom, so that many behaviors can yield goal-reaching rewards, but some of them are unnatural, do not perform well for all initial object states due to suboptimal contact modes, and thus correspond to shallower local minima of policy optimization.

\vspace{-0.2cm}
\subsection{Quantitative Evaluation with Estimation Error}
\label{sec:estimation_error}
\vspace{-0.2cm}
We evaluate the goal-reaching reward (as used in the previous section) of the policies under two types of state-estimation noise at varying magnitudes. The first is zero-mean Gaussian noise, similar to the domain randomization applied during training, which models random noise and occasional outlier observations. The second adds a constant mean shift to this baseline Gaussian noise throughout the episode; such systematic offsets are plausible in practice, arising, for example, from coordinate-frame misalignment between simulation and reality or from errors in camera-parameter estimation. We note that the per-axis noise standard deviation is set to 1 cm during training.

\begin{wrapfigure}{r}{0.65\linewidth}
    \centering
    \vspace{-0.4cm}
    \includegraphics[width=1.07\linewidth]{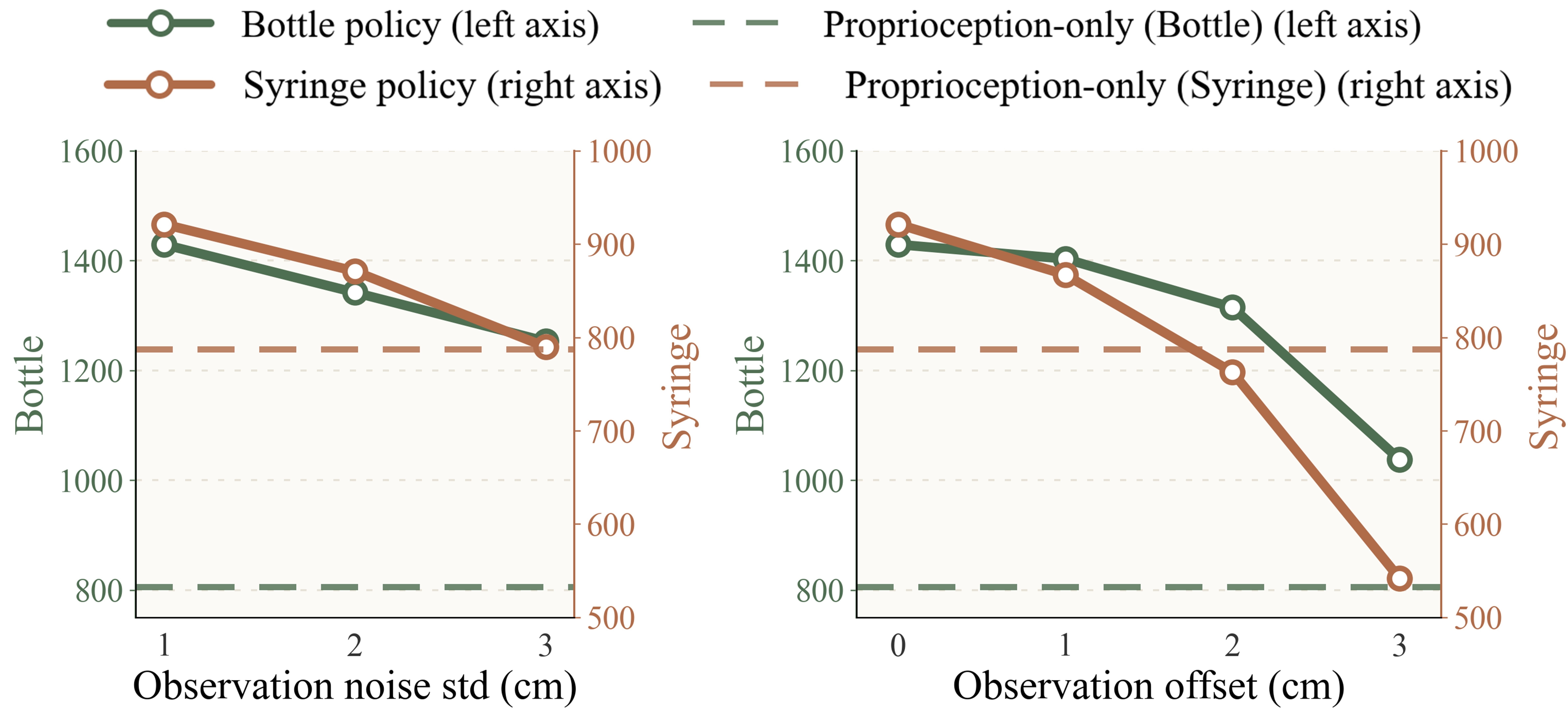}
    \caption{\textbf{Performance under different observation noise.}
    We report the goal-reaching reward under two noise types at varying
    magnitudes: zero-mean Gaussian noise (left), and the same Gaussian
    noise with a constant per-episode mean shift (right).}
    \label{fig:noise}
\end{wrapfigure}
Fig.~\ref{fig:noise} reports the results for each setting with proprioception-only baselines. For both tasks, performance remains above 60\% even under relatively strong noise (e.g., a 3 cm observation standard deviation), indicating considerable robustness to out-of-distribution observation noise. For the Syringe task, however, larger offset noise drives performance below the proprioception-only baseline. Given the performance of the proprioception-only policy, we conjecture that this robustness may stem from the policy's maintenance of proprioceptive history. Appendix~\ref{sec:marker_analysis} reports the Marker noise analysis, and Appendix~\ref{sec:physical_occlusion} evaluates pose-estimation errors under controlled physical occlusion.

\vspace{-0.2cm}
\subsection{Real-World System Capabilities with a Single Camera}
\begin{figure}[h]
    \centering
\includegraphics[width=0.9\linewidth]{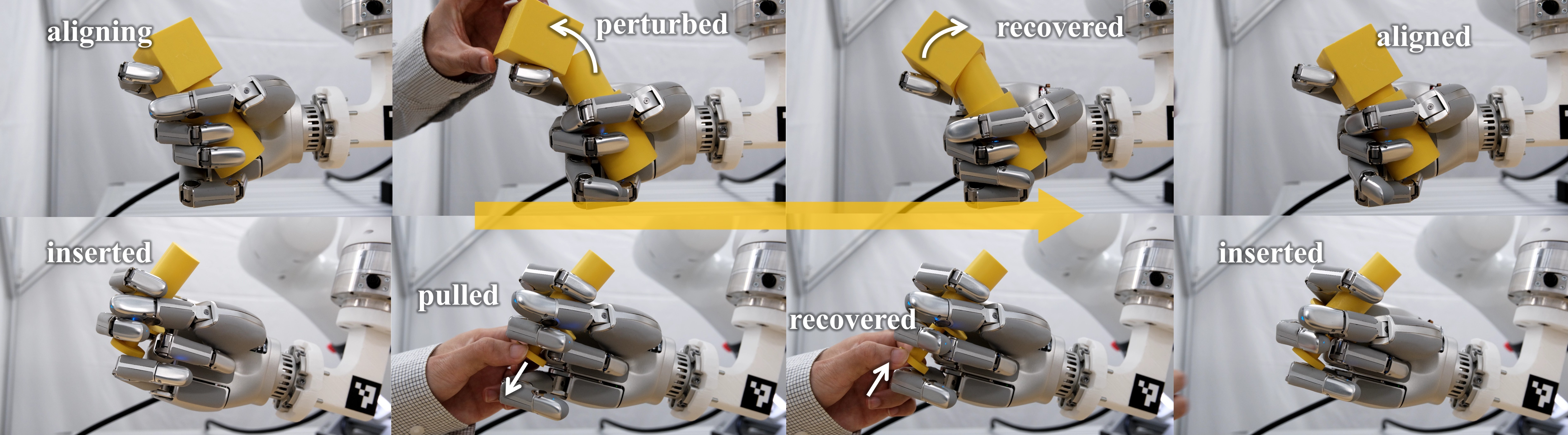}
    \caption{\textbf{Policy behaviors under unexpected perturbation.}}
    \label{fig:perturb}
    \vspace{-0.3cm}
\end{figure}

\vspace{-0.2cm}
We conduct real-world experiments for the three tasks using only a depth camera. During the tasks, the fingers perform subtle in-hand reorientation of the objects to aid alignment and insertion, and the pinky finger is often used to increase insertion depth, as shown in Fig. \ref{fig:teaser}.
\vspace{-0.15cm}
\begin{wraptable}{r}{0.7\linewidth}
    \centering
    \footnotesize
    \setlength{\tabcolsep}{3pt}
    \caption{\textbf{Closed-loop control versus open-loop replay.} Number of successful trials (out of 20) under the alignment and assembly criteria.}
    \label{tab:closedloop}
    \begin{tabular}{lccccccc}
        \toprule
        \multirow{2}{*}{Method} & \multicolumn{2}{c}{Bottle} & \multicolumn{2}{c}{Syringe} & \multicolumn{2}{c}{Marker} \\
        \cmidrule(lr){2-3} \cmidrule(lr){4-5} \cmidrule(lr){6-7}
         & Align & Assembly & Align & Assembly & Align & Assembly  \\
        \midrule
        Closed-loop (ours) & 18 & 15 & 18 & 17 & 16 & 16 \\
        Open-loop replay   & 2 & 2 & 1 & 0 & 0 & 0 \\
        \bottomrule
    \end{tabular}
    \vspace{-0.3cm}
\end{wraptable}
\vspace{-0.4cm}
\paragraph{Closed-loop control versus open-loop replay.}
To verify that closed-loop feedback is necessary, we run 20 consecutive trials comparing our closed-loop policy against open-loop replay of successful simulation trajectories. We record whether the \emph{alignment} criterion (insertion depth 
over 1\,cm) and the \emph{assembly} criterion (final depth within 1\,cm of the target) are met, as shown in Table \ref{tab:closedloop}. Open-loop replay achieves successful alignment only three times. Without feedback, the action sequences cannot compensate for sim-to-real dynamics, so early deviations accumulate, and the parts fail to align. The closed-loop policy succeeds far more often. A main failure case in the Bottle and Marker tasks is that an unfavorable initial cap pose prevents a stable pinch, leaving the cap tilted at an unrecoverable angle or dropped. In the Bottle and Syringe tasks, certain fingers can obstruct the insertion path, a state the policy has not learned to clear. \vspace{-0.2cm}
\paragraph{Robustness against Perturbation.}
We evaluate whether the policy has the ability to recover from unexpected deviations during the assembly process by artificially altering the state of one of the manipulated objects. Fig. \ref{fig:perturb} illustrates the policy’s recovery behaviors. In the Bottle task, the policy controls the index finger to correct an artificially induced misalignment of the bottle cap and realigns it without a pinch grasp. In the Syringe task, we resist the hand’s grasping force to pull out the plunger, and the hand quickly restores it to the fully assembled state. These experiments validate that the policies have some robustness against unexpected state deviations.

\begin{figure}[h]
    \centering
\includegraphics[width=0.7\linewidth]{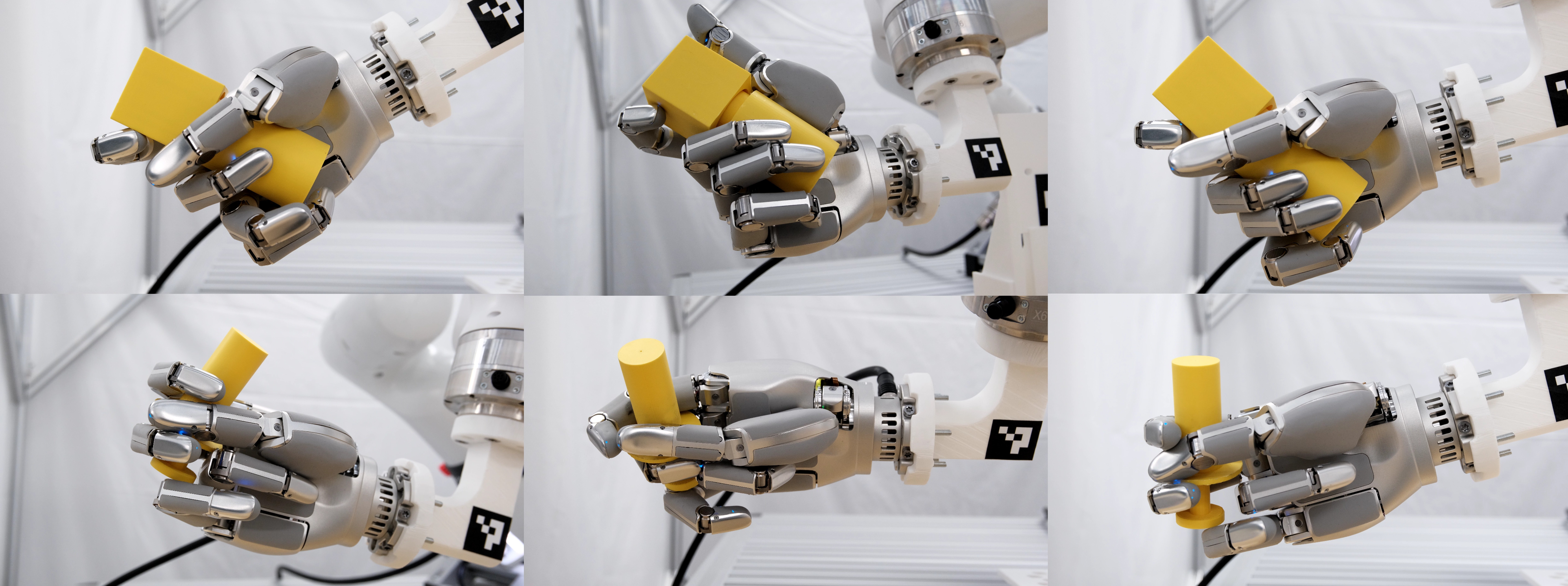}
    \caption{\textbf{In-hand assembly under different hand tilts.}}
    \label{fig:tilt}
    \vspace{-0.4cm}
\end{figure}

\paragraph{Assembly under Various Hand Tilts.}
In Fig. \ref{fig:tilt}, we present physical experiments demonstrating assembly with a single policy trained across various wrist-tilt angles, where the wrist pose is additionally provided as a policy observation. The policy succeeds across a range of tilt angles, showing that a single policy can adapt its finger coordination to the changing relative direction of gravity. However, this experiment reveals a limitation of our method: current rigid-body simulators~\cite{physx} cannot model planar (patch) contacts, so a pinch grasp degenerates into two contact points and becomes unstable in simulation~\cite{ye2025power}. This makes it hard for the policy to reliably pinch-control $\heldobj$ at certain tilt angles, e.g., when the object's long axis starts near horizontal. Soft finger pads alleviate the instability in the real world, but introduce a sim-to-real gap. Consequently, the policy cannot learn decisive pinch behaviors in these configurations during simulation, leading to failures in real-world deployment.

\vspace{-0.2cm}
\section{Limitations and Future Work}
\label{sec:limitations}
\vspace{-0.2cm}
A first limitation is that we study only a fragment of the in-hand manipulation workflow, rather than the full pipeline from grasping multiple objects off the table and adjusting their orientation for the subsequent assembly. The current isolation of in-hand manipulation requires artificial supporting force in some experiments until the policies engage, which may introduce factors that are difficult to model or quantify. However, a full workflow becomes long-horizon and involves skills, including multi-object grasping~\cite{seqmultigrasp,seqgrasp} and reorienting objects from their initial placement to the desired in-hand pose~\cite{chen2023sequential}, which remain open challenges. Second, we only currently include three tasks. In the future, we aim to implement more general part geometries and assembly types, such as threading and friction-fit, in the current suite to expand this work into a systematic benchmark for hardware and algorithms of in-hand manipulation. 

\vspace{-0.2cm}
\section{Conclusion}
\label{sec:conclusion}
\vspace{-0.2cm}

We introduced in-hand assembly, where a single dexterous hand must both stabilize and assemble objects without external fixtures or a second arm. With its requirement for intricate contact transitions and manipulation precision, this kind of task can serve as a benchmark for machine dexterity. We propose a principled RL solution with lightweight reward design, which distinguishes the roles between fingers and differentiates this RL task from other in-hand manipulation tasks. We validate its real-world effectiveness across three different tasks. These results suggest that anthropomorphic dexterity is not merely a matter of many degrees of freedom, but can be realized through human-like functional finger specialization and coordinated contact use, moving robotic hands toward human-level multi-functional dexterous manipulation.

\clearpage
\bibliography{references}

\clearpage

\appendix
\section{Training Implementation}
\label{sec:appendxA}
In this section, we provide a detailed overview of the training setup omitted from the main text, including simulation setup, implementation details for object initialization, policy representation, RL algorithm hyperparameters, and domain randomization.

\paragraph{Simulation setup.} 
We use IsaacSim \cite{isaacsim} for simulation-based training. We set up the simulator for efficient contact-rich simulation largely following practices in \cite{narang2022factory}, which is a PhysX-based \cite{physx} implementation for simulating assembly tasks. In particular, we use signed distance function (SDF)-based collisions for the meshes in simulation; the contacts are generated with a contact patch for number reduction and solved with a Gauss-Seidel solver.

The physics timestep is $1/120\,\mathrm{s}$, and the scene uses 16 position iterations and one velocity iteration for rigid-object contacts.  The fixed-base hand articulation uses a larger position-iteration budget, 32 position iterations, and one velocity iteration, which improves stability during multi-finger contact. For task objects, the contact offset is $5\,\mathrm{mm}$ and the rest offset is zero.  The friction offset threshold is $10\,\mathrm{mm}$ and the friction correlation distance is $6.25\,\mathrm{mm}$, which control friction patch generation and merging. 

\paragraph{Initial object state construction implementation}
Instead of directly teleporting the objects to randomized poses that can cause interpenetration with the hand, we apply a short reset-only wrench perturbation.  The procedure starts from the nominal state: the hand is written to its reset joint configuration, $\fixedobj$ and $\heldobj$ are placed at their task-specific poses extracted from the reference snapshot, as described in \S \ref{sec:method:init}. For each asset, we set a bounded local-frame translational displacement $\Delta p$ and a bounded local-frame rotation vector $\Delta\theta$, which represent expected maximum pose perturbation around the nominal object states.  The perturbation is not applied by directly overwriting the object pose.  Instead, it is converted into an external force and torque over a short reset window of $N=2$ physics steps. With simulator timestep $\Delta t$, the implementation uses \[ c = \frac{2}{\Delta t^2 N(N+1)} \] as the acceleration coefficient for the reset perturbation.  For an object with mass $m$, the translational force is \[ F^W = R(\bar{q})\,\Delta p \; m c, \] where $R(\bar{q})$ maps the sampled local displacement into the world frame. The rotational perturbation uses an approximate diagonal inertia model for the object to be manipulated.  If $I$ is the diagonal inertia estimate, the local torque is \[ \tau^L = I\,\Delta\theta\,c, \] and it is then rotated into the world frame before being applied to the object. During the reset-only wrench rollout, the robot is repeatedly rewritten to its reset joint state and root state. Thus, the hand acts as a static collision body while the objects settle under the sampled perturbation.  After the short simulation window, the external wrench is cleared.  The resulting object poses are clamped relative to their nominal states so that neither translation nor rotation exceeds the configured bounds.  Finally, both object root velocities are set to zero.

\paragraph{Termination condition.} Each task terminates after a certain number of steps. We use task-dependent episode horizons in simulation, where the Bottle task's episode length is 20s, the Syringe's is 10s, and the Marker's is 20s. Early termination occurs when one object moves 0.5m away from the hand, e.g., falling from the hand.

\paragraph{Policy representation.}
The policy is a recurrent Gaussian whose backbone is
a two-layer LayerNorm-LSTM with 1024 hidden units,
followed by a $[512, 128, 64]$ ELU MLP that emits the action mean. A separate head shares the same backbone to estimate the discounted value. The history MLP baseline is implemented as an $[1024, 512, 128, 64]$ ELU MLP.

\paragraph{RL Setup.} Policies are trained with the Isaac Lab wrapper \cite{mittal2025isaaclab} for \texttt{rl\_games} \cite{rl-games2021}, where the Proximal Policy Optimization (PPO) \cite{schulman2017proximal} implementation is used. Task variants with randomized hand tilt use four times as many parallel environments as those with fixed hand tilt; their actor and central-value minibatch sizes are scaled by four compared to the fixed tilt setup to keep the per-update minibatch ratio comparable. The detailed hyperparameters are listed in Table \ref{tab:rl_params}.

\begin{table}[h]
  \centering
  \caption{Key RL training parameters.  Entries separated by ``/'' denote the fixed hand tilt task and randomized hand tilt settings, respectively.}
  \vspace{0.3cm}
  \label{tab:rl_params}
  \begin{tabular}{ll}
    \toprule
    Parameter & Value \\
    \midrule
    Parallel environments & 256 / 1024 \\
    Rollout horizon & 128 steps \\
    Samples per rollout & 32768 / 131072 \\
    PPO minibatch size & 2048 / 8192 \\
    Central critic minibatch size & 512 / 2048 \\
    Optimization epochs per rollout & 4 \\
    Recurrent sequence length & 128 \\
    Discount factor $\gamma$ & 0.995 \\
    GAE parameter $\lambda$ & 0.95 \\
    Learning rate & $1.0\times10^{-4}$ \\
    Learning-rate schedule & adaptive KL \\
    KL threshold & 0.008 \\
    PPO clip range & 0.2 \\
    Entropy coefficient & 0.0 \\
    Critic loss coefficient & 2.0 \\
    Gradient-norm clipping & 1.0 \\
    Bounds loss coefficient & $1.0\times10^{-4}$ \\
    Input/value normalization & enabled \\
    Mixed precision & enabled \\
    \bottomrule
  \end{tabular}
\end{table}

\paragraph{Domain randomization.}
We model observation errors, including those from state estimation and joint angle reading, and actuation noise by domain randomization. We apply a substantial degree of randomization to the properties of the object, such as mass and friction, while simultaneously introducing randomization into the properties of the manipulator, such as control stiffness and damping. Some variables are sampled once at reset and remain constant throughout the episode; these represent persistent physical uncertainties.  Other variables are resampled at every policy step; these represent sensing or command noise.  A third group is event-driven: the perturbation is triggered intermittently within an episode rather than being constant or continuously resampled.   Table~\ref{tab:dr} summarizes the components of domain randomization and their ranges.

\begin{table}[t]
    \centering
    \small
    \caption{Domain randomization applied during training. Per-step
    parameters are resampled at each policy step; per-episode parameters
    are fixed within an episode and resampled at reset.}
    \label{tab:dr}
    \begin{tabular}{lll}
        \toprule
        Parameter & Distribution & Cadence \\
        \midrule
        Joint-position observation noise   & $\sigma = 0.01$\,rad                       & per-step \\
        Joint-velocity observation noise   & $\sigma = 0.10$\,rad/s                     & per-step \\
        Object position observation noise  & $\sigma = 0.01$\,m                         & per-step \\
        Object orientation observation noise & $\sigma = 0.05$ (quat. components)       & per-step \\
        Action noise / delay               & $\sigma = 0.02$ / $\{0,1,2,3\}$ steps     & per-step \\
        Gravity magnitude / tilt           & $\mathcal{U}[9.51, 10.11]$\,m/s$^{2}$ / $\sigma = 0.05$\,rad & per-step bias \\
        Surface friction (per material)    & $\mathcal{U}[0.5, 2.0] \times$ nominal     & per-episode \\
        Object mass                        & $\mathcal{U}[0.05, 0.15]$\,kg (absolute)   & per-episode \\
        Joint stiffness / damping          & $\mathcal{U}[0.5, 2.0] \times$ USD value   & per-episode \\
        Restitution                        & $\mathcal{U}[0.0, 0.5]$                    & per-episode \\
        Center-of-mass offset              & $\mathcal{U}[\pm 0.01\,\text{m}]^3$        & per-episode (epoch $\ge 600$) \\
        Warmup steps              & $\mathcal{U}[7, 12] steps$        & per-episode \\
        External disturbance force
  & \shortstack[l]{random direction, $\mathcal{U}[0, 0.3]$ N,\\
                  interval $\mathcal{U}\{20,\ldots,60\}$}
  & event-driven \\

        \bottomrule
    \end{tabular}
\end{table}

\begin{figure}[h]
    \centering
\includegraphics[width=1\linewidth]{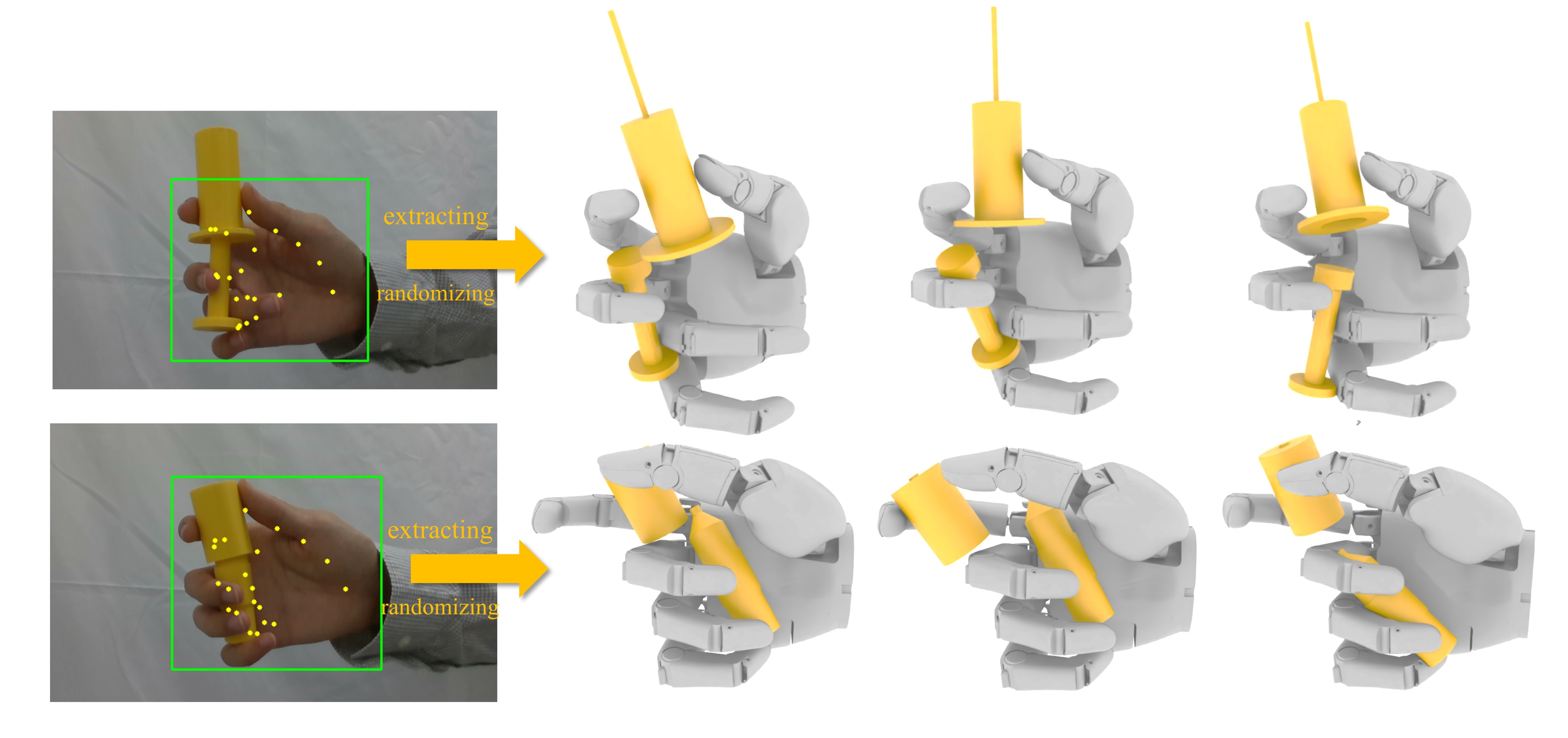}
    \caption{\textbf{Supplementary demonstration of state initialization from a human reference snapshot.} On the left is the human reference snapshot for the Bottle task; the three simulated states on the right are examples of random initial states generated based on the reference hand pose.}
    \label{fig:appx_init}
\end{figure}

\section{Real-World Deployment}
\label{sec:method:deploy}
\paragraph{Robotic hand deployment procedure.}
In our experiments, we first fixed the wrist of the robotic hand and set the task-specific initial joint angles described in \S \ref{sec:method:init}, then manually placed $\fixedobj$ between the middle, ring, and pinky fingers, and placed $\heldobj$ between the index finger and the thumb. Since these two static fingers often cannot provide a stable pinch grip, the human continues to hold the object still until neural network control is activated, as shown in the first row of Fig. \ref{fig:teaser}. In our future work, we plan to develop a fully automated process (starting with object grasping), as mentioned in \S \ref{sec:limitations}, to address the issue of objects needing to be manually positioned at the start.

During autonomous manipulation, joint angles (proprioception) are obtained directly from the hand SDK at 500 Hz and decimated to the control rate (15 Hz). Policy actions are clipped, slew-limited to $0.12$ rad per step, and interpolated to 60 Hz before the hand's position controller. A tracking-error-driven stall-relax loop monitors per-joint commanded vs. measured position and slightly relaxes targets that exceed a stall threshold, keeping motor currents below thermal limits during long, contact-rich grasps. Otherwise, jamming between objects caused by manipulating two objects simultaneously would lead to rapid overheating of the robotic hand and damage to the hardware.

\begin{figure}[h]
    \centering
\includegraphics[width=0.5\linewidth]{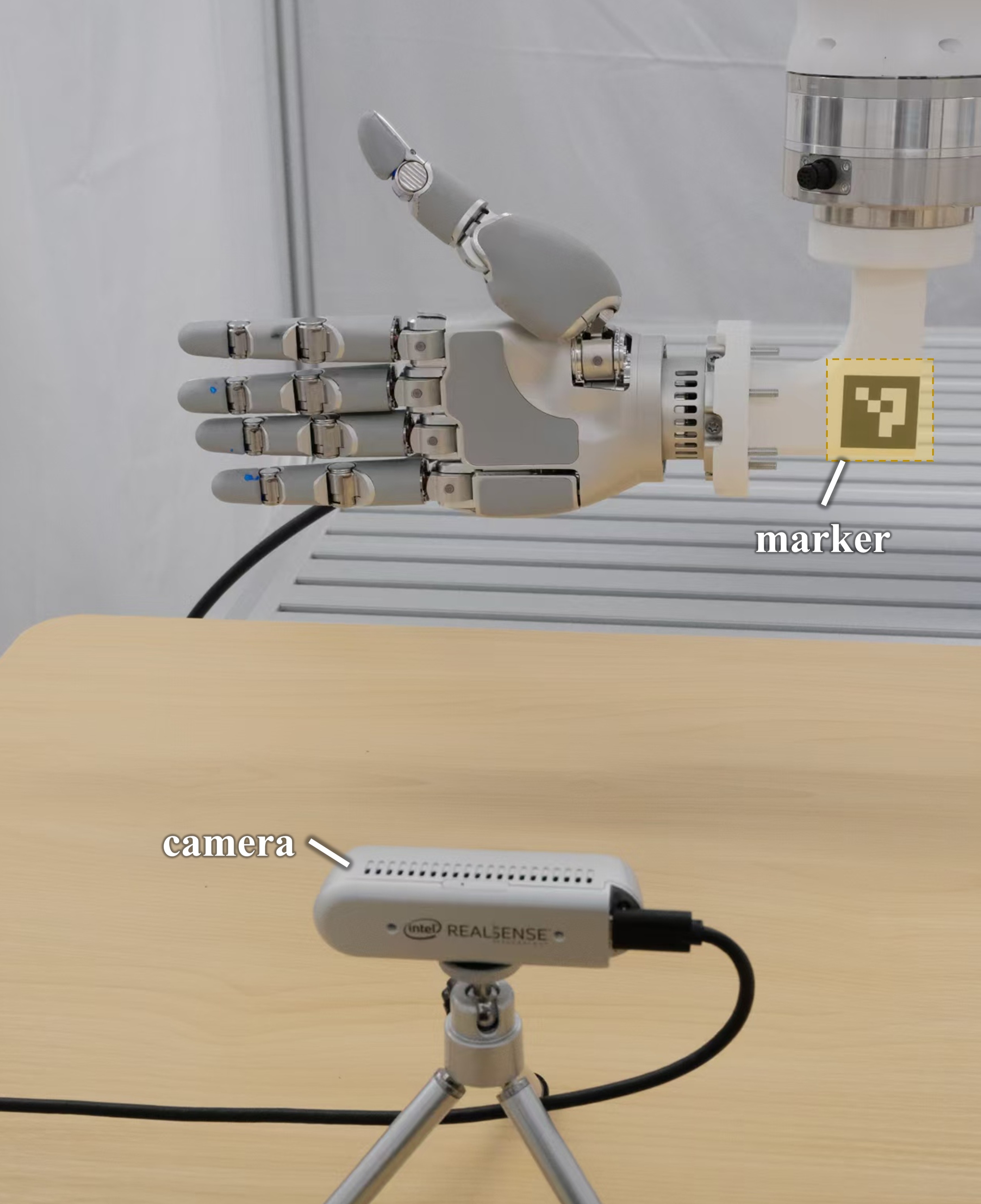}
    \caption{\textbf{The experiment scenario.}}
    \label{fig:appx_b}
\end{figure}

\paragraph{Camera calibration.}
For real-world deployment, we decouple the persistent hand-fixture geometry from the per-run camera extrinsics.  The fiducial marker, as shown in Fig. \ref{fig:appx_b}, is mounted on the hand bracket rather than placed at the task-frame origin.  At the beginning of each run, the RGB-D camera observes this marker and estimates $T_{\mathrm{marker}\leftarrow\mathrm{cam}}$ over a short calibration window of 10 frames.  We use sub-pixel ArUco corner refinement, the planar-marker PnP solution, and robust aggregation of the detected marker poses to reduce frame-level corner noise.  The live marker-camera transform is then composed with a bracket calibration exported from the CAD model, $T_{\mathrm{base}\leftarrow\mathrm{marker}}$, and the simulator task-frame transform, $T_{\mathrm{task}\leftarrow\mathrm{base}}$:
\[
T_{\mathrm{task}\leftarrow\mathrm{obj}} =
T_{\mathrm{task}\leftarrow\mathrm{base}}\,
T_{\mathrm{base}\leftarrow\mathrm{marker}}\,
T_{\mathrm{marker}\leftarrow\mathrm{cam}}\,
T_{\mathrm{cam}\leftarrow\mathrm{obj}} .
\]
The policy therefore receives object poses in the same task-frame convention as simulation, while the camera may be repositioned between trials as long as the bracket marker is visible during startup.  At the same startup stage, when policies trained with variable wrist pose are used, a second reference marker on the table is used to infer the current wrist orientation.

\paragraph{Pose-tracking stabilization.}
Object tracking runs on a 640$\times$480, 30\,Hz RGB-D stream.  The two object meshes are first registered from color-segmented masks, with 10 refinement iterations at initialization and 2 refinement iterations in subsequent tracking steps.  Since our objects are close to rotationally symmetric, we initialize the mesh orientation with a consistent upright convention and pass to the policy only the centroid and the body $z$-axis, rather than relying on the poorly constrained yaw angle.  Stability during manipulation is enforced by rejecting physically implausible updates before they reach the policy.  First, the RealSense depth stream is spatially and temporally filtered.  Second, a depth-consistency gate rejects a tracked pose if the predicted object-center depth differs from the measured depth by more than 8\,cm, which commonly occurs when a finger occludes the object and the tracker begins to explain the hand surface.  Third, a frame-to-frame translation jump larger than 6\,cm is rejected, and the internal tracker state is rolled back to the last accepted pose.  Accepted poses are then smoothed with an exponential filter, using linear interpolation for translation and spherical interpolation for rotation with a coefficient of 0.7.  If low-confidence tracking persists for 15 frames, the object is re-registered from color/depth contours, with contour assignment chosen by proximity to the last projected object centers to reduce identity swaps.  Finally, the downstream policy consumes an atomically written latest-pose message in the task frame; transient invalid estimates are held at the most recent accepted value for at most 120 frames and marked with reduced confidence, preventing momentary occlusions from becoming discontinuous observation dropouts.

\clearpage
\section{Marker Ablation and Noise Analyses}
\label{sec:marker_analysis}
\begin{figure}[htbp]
    \centering
    \includegraphics[width=\statfigwidth]{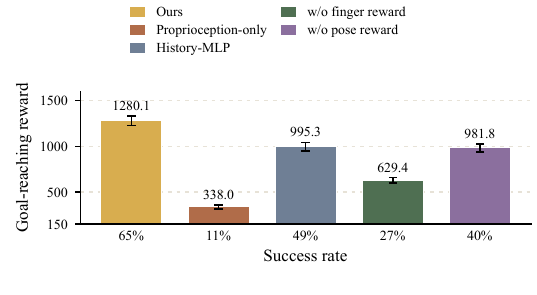}
    \caption{\textbf{Marker ablation results.} Bars show goal-reaching rewards for the same variants as in Figure~\ref{fig:ablation}. Reward values are annotated above the bars; assembly success rates over 500 episodes are shown below them.}
    \label{fig:marker_ablation}
\end{figure}

We extend the ablation study in Section~\ref{sec:ablation} and the noise analysis in Section~\ref{sec:estimation_error} to the Marker task. Figure~\ref{fig:marker_ablation} reports the goal-reaching reward for the full policy and the same four ablations used in the main text. We also report assembly success rates over 500 episodes per variant under domain randomization, using the same success criterion as in the real-world experiments. The full policy achieves a goal-reaching reward of 1280.1 and a success rate of 65\%, compared with 338.0 and 11\% for the proprioception-only policy. Removing the finger-function reward or the reference-pose reward also reduces performance.

\begin{figure}[htbp]
    \centering
    \includegraphics[width=0.6955\textwidth]{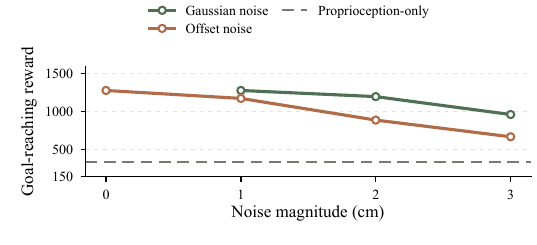}
    \caption{\textbf{Marker performance under observation noise.} The horizontal axis denotes the Gaussian noise standard deviation or the constant offset magnitude, respectively, using the noise definitions in Section~\ref{sec:estimation_error}. The dashed line indicates the proprioception-only baseline.}
    \label{fig:marker_noise}
\end{figure}

Figure~\ref{fig:marker_noise} evaluates the Marker policy under the same Gaussian and offset noise settings as in Figure~\ref{fig:noise}. Performance remains stable under moderate noise and decreases as the noise magnitude increases. At a Gaussian noise standard deviation of 3\,cm, the goal-reaching reward is 963.7; with a constant offset of 3\,cm added to the baseline Gaussian noise, it is 670.9. Both exceed the proprioception-only baseline of 338.0. These results support robustness to moderate pose-estimation errors, while also showing degradation under larger errors.

\clearpage
\section{Noise Model and Physical Occlusion}
\label{sec:physical_occlusion}
Gaussian noise models random pose-estimation errors, including occasional outliers, whereas constant offsets model systematic calibration errors. Neither directly models physical occlusion. To examine estimation errors under occlusion, we keep each object stationary, vary its visible fraction, and measure the translation error of FoundationPose relative to an unoccluded reference.

\begin{figure}[htbp]
    \centering
    \includegraphics[width=\textwidth]{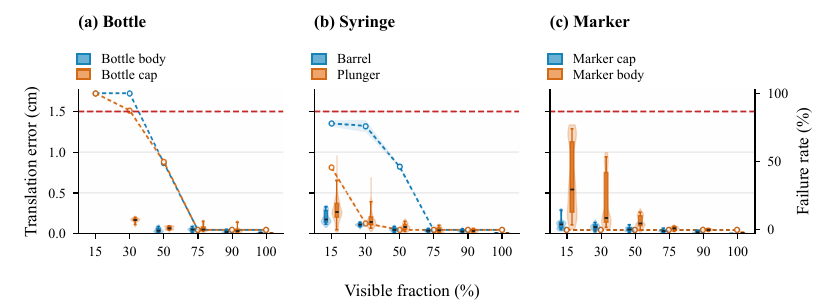}
    \caption{\textbf{Pose-estimation errors under controlled physical occlusion.} Each object remains stationary while its visible fraction varies. Blue and orange distributions show translation errors of valid estimates relative to an unoccluded reference for the two parts in each task. Dashed curves show the corresponding failure rates on the right axis; the red dashed line marks 1.5\,cm.}
    \label{fig:physical_occlusion}
\end{figure}

Figure~\ref{fig:physical_occlusion} reports the error distributions and failure rates for both parts in each task. Most valid estimates differ from the unoccluded reference by less than 1.5\,cm. However, severe occlusion can produce large errors or tracking loss. A translation error greater than 2\,cm or tracking loss is counted as a failure; these failures are reported separately from the distributions of valid estimates. Thus, the small errors among valid estimates do not imply reliable tracking at all visibility levels. The Gaussian and offset noise experiments assess policy robustness to pose errors, while this controlled-occlusion experiment characterizes a physical source of those errors.

\end{document}